\documentclass[sigconf, nonacm]{acmart}

\setcopyright{none}
\renewcommand\footnotetextcopyrightpermission[1]{}
\usepackage{enumitem}
\usepackage{pifont}
\usepackage[normalem]{ulem}

\AtBeginDocument{%
  }

\newcommand{\pname}[1]{QCFuse{#1}}

\usepackage[ruled,vlined,linesnumbered]{algorithm2e}
\DontPrintSemicolon
\SetKwInOut{Input}{Input}
\SetKwInOut{Output}{Output}

\begin{document}
\setlist[itemize]{left=0pt, topsep=0pt, itemsep=0pt, parsep=0pt}
\setlist[enumerate]{left=0pt, topsep=0pt, itemsep=0pt, parsep=0pt}

\title{QCFuse: Query-Aware Cache Fusion via Compressed View for Efficient RAG Serving}

\newcommand{\affmark}[1]{\textsuperscript{#1}}
\newcommand{\corrmark}{\textsuperscript{*}}

\makeatletter
\def\@mkauthors{%
  \global\setbox\mktitle@bx=\vbox{%
    \noindent\box\mktitle@bx\par\medskip
    \centering
    {\normalfont\large
    Jianxin Yan\affmark{1}, Wangze Ni\affmark{1}\corrmark, Zhenxin Li\affmark{1}, Jiabao Jin\affmark{2}, Zhitao Shen\affmark{2},\\
    Haoyang Li\affmark{3}, Jia Zhu\affmark{4}, Peng Cheng\affmark{5}, Xuemin Lin\affmark{6}, Lei Chen\affmark{7,8}, Kui Ren\affmark{1}\par}
    \vskip 0.3em
    {\normalfont\normalsize
    \affmark{1}Zhejiang University, Hangzhou, China \quad
    \affmark{2}Ant Group, Shanghai, China \\
    \affmark{3}The Hong Kong Polytechnic University, Hong Kong, China \quad
    \affmark{4}Zhejiang Normal University, Jinhua, China \\
    \affmark{5}Tongji University, Shanghai, China \quad
    \affmark{6}The Chinese University of Hong Kong, Shenzhen, China \\
    \affmark{7}The Hong Kong University of Science and Technology (Guangzhou), Guangzhou, China\\
    \affmark{8}The Hong Kong University of Science and Technology, Hong Kong, China\par}
    \vskip 0.3em
    {\normalfont\normalsize
    \texttt{\{yanjianxin,niwangze,zhenxin,kuiren\}@zju.edu.cn}, \nolinkurl{jiabaojin@stu.ecnu.edu.cn},\\
    \nolinkurl{zhitao.szt@antgroup.com}, \nolinkurl{haoyang-comp.li@polyu.edu.hk}, \nolinkurl{jiazhu@zjnu.edu.cn},\\
    \nolinkurl{cspcheng@tongji.edu.cn},
    \nolinkurl{xuemin.lin@gmail.com}, \nolinkurl{leichen@cse.ust.hk}\par}
    \medskip
  }%
}
\makeatother


\author{Jianxin Yan}
\affiliation{%
  \institution{Zhejiang University}
  \city{Hangzhou}
  \state{China}
}
\email{yanjianxin@zju.edu.cn}

\author{Wangze Ni}
\affiliation{%
  \institution{Zhejiang University}
  \city{Hangzhou}
  \state{China}
}
\email{niwangze@zju.edu.cn}

\author{Zhenxin Li}
\affiliation{%
  \institution{Zhejiang University}
  \city{Hangzhou}
  \state{China}
}
\email{zhenxin@zju.edu.cn}

\author{Jiabao Jin}
\affiliation{%
  \institution{East China Normal University}
  \city{Shanghai}
  \state{China}
}
\email{jiabaojin@stu.ecnu.edu.cn}

\author{Zhitao Shen}
\affiliation{%
  \institution{Ant Group}
  \city{Shanghai}
  \state{China}
}
\email{zhitao.szt@antgroup.com}

\author{Haoyang Li}
\affiliation{%
  \institution{The Hong Kong Polytechnic University}
  \city{Hong Kong}
  \country{China}
}
\email{haoyang-comp.li@polyu.edu.hk}

\author{Jia Zhu}
\affiliation{%
  \institution{Zhejiang Normal University}
  \city{Jinhua}
  \country{China}
}
\email{jiazhu@zjnu.edu.cn}

\author{Peng Cheng}
\affiliation{%
  \institution{Tongji University}
  \city{Shanghai}
  \country{China}
}
\email{cspcheng@tongji.edu.cn}

\author{Xuemin Lin}
\affiliation{%
  \institution{The Chinese University of Hong Kong, Shenzhen}
  \city{Shenzhen}
  \country{China}
}
\email{xuemin.lin@gmail.com}

\author{Lei Chen}
\affiliation{%
  \institution{The Hong Kong University of Science and Technology (Guangzhou)}
  \city{Guangzhou}
  \country{China}
}
\affiliation{%
  \institution{The Hong Kong University of Science and Technology}
  \city{Hong Kong}
  \country{China}
}
\email{leichen@cse.ust.hk}

\author{Kui Ren}
\affiliation{%
  \institution{Zhejiang University}
  \city{Hangzhou}
  \state{China}
}
\email{kuiren@zju.edu.cn}

\renewcommand{\shortauthors}{Yan et al.}

\begin{abstract}
Retrieval-augmented generation (RAG) improves LLM answer quality by grounding generation in external evidence, but processing retrieved contexts makes the prefill stage a dominant serving cost. RAG cache fusion reduces this cost by reusing precomputed key-value (KV) caches for retrieved chunks and selectively recomputing tokens under the current prompt. Existing selectors, however, face a dilemma between quality and efficiency: fast query-agnostic or final-layer query-to-context selectors can miss request-relevant evidence, whereas full-view query-aware selectors require broad context and layer visibility before recomputation and therefore stall the layer-wise cache-fusion pipeline. We present \pname{}, a compressed-view query-aware selector for RAG cache fusion. \pname{} uses chunk-anchor query probing to condition user-query states on compact per-chunk anchors and critical-layer profiling to identify recomputation tokens without all-layer inspection. We implement \pname{} in SGLang and evaluate it on four open-weight LLMs across six datasets. \pname{} reaches full-prefill-level quality. At matched quality, \pname{} achieves an average prefill-time speedup of 1.7$\times$ over full prefill and 1.5$\times$ over ProphetKV, the strongest quality-preserving baseline.
\end{abstract}

\maketitle

\begingroup
\renewcommand\thefootnote{}\footnote{\noindent
Work done while Jianxin Yan and Zhenxin Li were interns at Ant Group.\\
\noindent *Corresponding Author. \\
\noindent The source code is available at \url{https://github.com/uYanJX/QCFuse}
}\addtocounter{footnote}{-1}
\endgroup

\section{Introduction}

\begin{figure}[t]
  \centering
  \includegraphics[width=\linewidth]{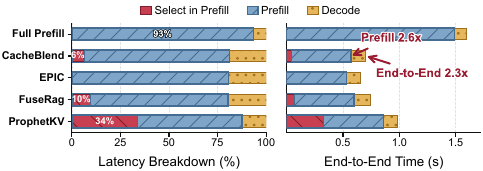}
  \caption{Time breakdown under RAG serving on Qwen3-8B. Cache-fusion methods use a 30\% recomputation ratio.}
  \Description{End-to-end time percentage breakdown on Qwen3-8B for MuSiQue and 2WikiMQA. The plot compares full prefill, direct reuse, and selective-recomputation methods; cache-fusion methods use a 30 percent recomputation ratio. It shows the share of total response time spent in each serving stage.}
  \label{fig:intro_e2e_breakdown}
\end{figure}

RAG is emerging as a data management paradigm that retrieves query-relevant evidence from trusted and up-to-date external data collections to help LLMs generate more accurate and reliable query responses in data-intensive applications~\cite{zhao2024chat2data,li2024llmdata,balaka2025pneuma,madden2024databases}.
The retrieved evidence and query are jointly fed into LLMs, which prefill the input to produce the corresponding key-value (KV) matrices for subsequent query response generation. Since the runtime of the prefill stage dominates the total time of query response generation in RAG scenarios (as shown in Figure~\ref{fig:intro_e2e_breakdown}), improving prefill efficiency has attracted growing attention from academia (e.g., the database community~\cite{Wang2026FromPC,gao2025apt,li2025hotprefix}) and industry (e.g., NVIDIA~\cite{nvidia2024tensorrtllm_chunked_prefill}).
A promising solution is \emph{KV cache fusion}~\cite{yao2025cacheblend,Wang2026FromPC,gim2024prompt,agarwal2025cache}, which selectively reuses some previous data's KV caches to alleviate prefill overhead. As shown in Figure~\ref{fig:intro_e2e_breakdown}, CacheBlend~\cite{yao2025cacheblend} accelerates the prefill stage by 2.6$\times$ and achieves a 2.3$\times$ end-to-end speedup over full prefill.

\begin{figure}[t]
  \centering
  \includegraphics[width=\linewidth]{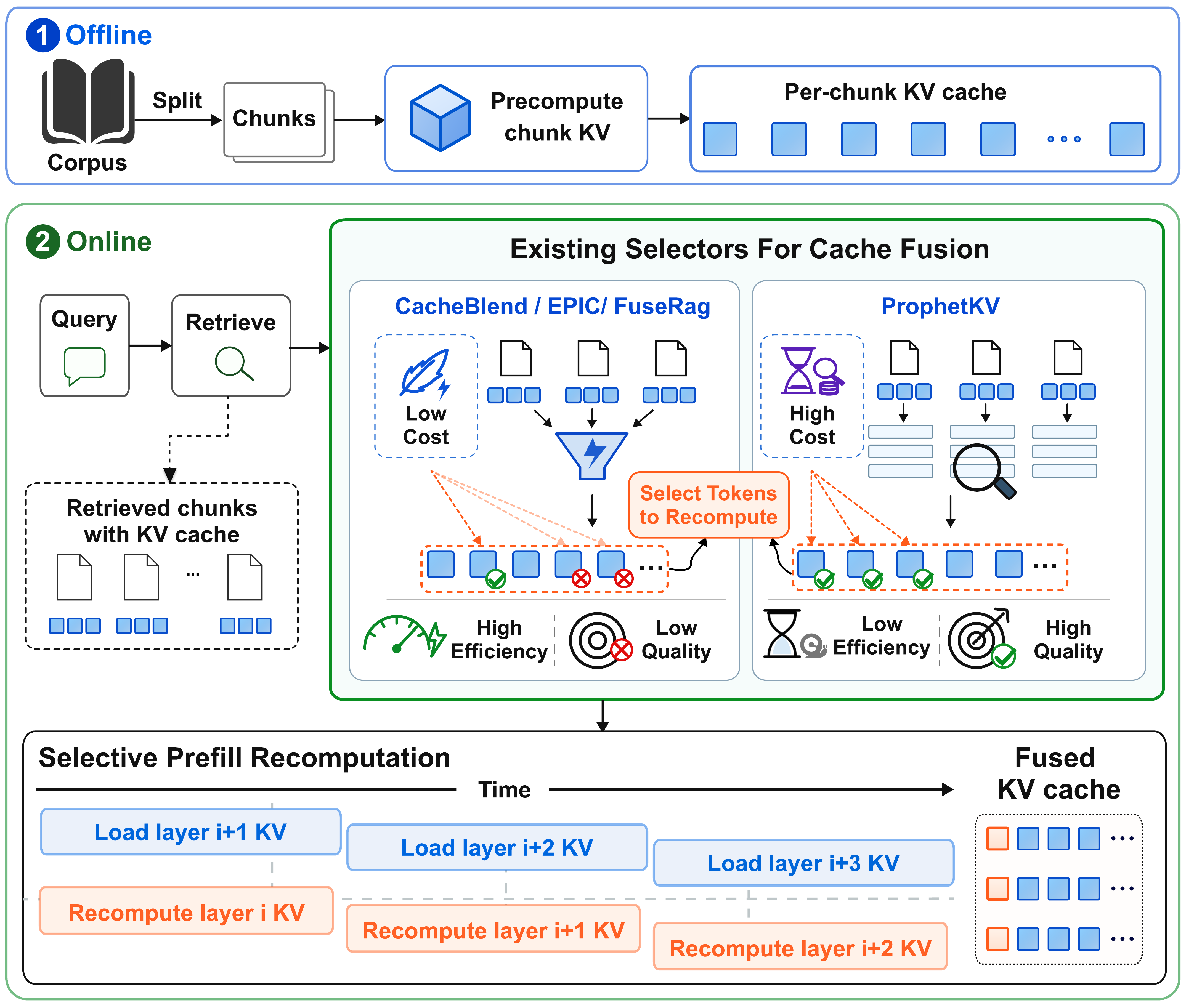}
  \caption{Cache-fusion workflow for RAG.}
  \label{fig:intro_blend_pipeline}
\end{figure}

KV cache fusion for RAG exploits a key workload property: while user queries vary across requests, the external RAG corpus is relatively stable.
As shown in the upper offline part of Figure~\ref{fig:intro_blend_pipeline}, the system partitions this corpus into reusable chunks, i.e., short evidence passages used as retrieval units, and stores their precomputed KV caches as per-chunk KV caches.
During online inference, as shown in the lower online part, the system retrieves chunks for the current query and loads their precomputed KV caches from storage to GPUs instead of reprocessing the chunks from scratch.
Because these chunk caches are computed independently and lack cross-chunk context, cache fusion selects a subset of context tokens for recomputation under the current context prompt to recover cross-chunk dependencies.
To enhance inference efficiency, cache fusion is organized as a layer-wise cache-fusion pipeline.
This pipeline overlaps KV-cache loading for subsequent layers with selective recomputation at the current layer, reducing prefill-stage overhead.

Although KV cache fusion has substantial potential to accelerate long-context generation, existing cache fusion systems still face a dilemma between \textbf{generation quality and efficiency}.
Since generation quality largely depends on which tokens are selected for recomputation, existing selectors can be grouped by how they incorporate the user query.
Firstly, query-agnostic selectors such as CacheBlend~\cite{yao2025cacheblend} and EPIC~\cite{hu2024epic} rely on static or context-only signals.
Secondly, FusionRAG~\cite{Wang2026FromPC} uses final-layer user-query-to-context attention as a lightweight query-aware signal.
These methods keep selection fast and can recover part of the quality lost by direct reuse.
On complex multi-hop tasks, they may miss key context tokens needed to answer the current user query, leaving a gap in full-prefill quality.
In contrast, ProphetKV~\cite{Wang2026ProphetKVUS} aggregates user-query-to-context relevance across chunks and layers, improving selection quality but requiring broad KV-cache visibility before recomputation, which stalls the layer-wise cache-fusion pipeline.
As shown in Figure~\ref{fig:intro_e2e_breakdown}, at a 30\% recomputation ratio, ProphetKV's selection stage accounts for 34\% of end-to-end time.
This gap raises a central question: \emph{how can cache fusion use query-aware token selection to preserve generation quality without sacrificing speed?}

However, answering this question is challenging because accurately identifying request-relevant tokens for recomputation requires comparing the user query with the retrieved evidence.
An accurate query-aware selector therefore needs two evidence views: how the current user query relates to the retrieved context, and which model-layer attention signals are useful for localization.
A full-view selector obtains these views by exposing the query probe to the complete retrieved context and broad layer-wise attention signals.
These views require KV-cache transfers before recomputation starts.
The layer-wise cache-fusion pipeline would otherwise overlap such transfers, but full-view selection turns them into serialized pre-fusion work.
This creates two challenges for the selector:
\begin{itemize}
    \item \textbf{Challenge 1: Token-view conditioning bottleneck.} Query-aware selection needs user-query states conditioned on retrieved evidence. A query-only probe misses evidence relevance, while full-context probing requires broad context visibility before the pipeline can start.
    \item \textbf{Challenge 2: Layer-view localization bottleneck.} Token localization depends on layer-wise attention signals. Final-layer-only signals can be noisy, while all-layer analysis requires loading context KV across layers and stalls the pipeline.
\end{itemize}

To address these challenges, we present \pname{}, whose key idea is to retain query-aware selection of tokens to recompute while compressing the evidence exposed to the selector:
\begin{itemize}
    \item To address Challenge 1, \pname{} introduces \textbf{chunk-anchor query probing} to compact the token view. During the offline stage, \pname{} selects representative anchor tokens within each reusable corpus chunk to build compact per-chunk anchor sets. During online serving, \pname{} probes the user query over these chunk anchors and ranks original context tokens for recomputation without a full-context probing pass (Section~\ref{sec:anchor_context}).
    \item To address Challenge 2, \pname{} performs \textbf{critical-layer profiling} to compact the layer view. During the offline stage, \pname{} uses a model-level diagnostic to identify critical layers whose user-query-to-context attention best localizes query-relevant context tokens. During online serving, the selector examines only these critical-layer signals and selects original context tokens for recomputation, avoiding an all-layer scan (Section~\ref{sec:critical_layer}).
\end{itemize}

We evaluate \pname{} across multiple open-weight LLMs, datasets, and cache-fusion baselines, showing that compressed-view selection reaches full-prefill-level quality while improving serving efficiency. At matched quality, \pname{} achieves an average prefill-time speedup of 1.7$\times$ over full prefill and 1.5$\times$ over ProphetKV, the strongest quality-preserving baseline. In summary, this paper makes the following contributions:
\begin{itemize}
    \item We identify token-view and layer-view evidence bottlenecks for query-aware selective recomputation in layer-wise cache fusion (Section~\ref{sec:query_selection}).
    \item We introduce a \textbf{compressed-view selector} that combines chunk-anchor query probing with model-specific critical-layer profiling to reduce pre-fusion KV-cache loading (Sections~\ref{sec:anchor_context} and~\ref{sec:critical_layer}).
    \item We implement \pname{} as a pipelined cache-fusion system in SGLang and integrate a Triton-optimized selective KV-cache recomputation path (Section~\ref{sec:qcfuse_cache_pipeline}).
    \item We provide a broad empirical study showing that the quality--latency gains hold across models and benchmarks (Section~\ref{sec:exp}).
\end{itemize}

The rest of this paper is organized as follows. Section~\ref{sec:prelim} gives the background on RAG KV cache fusion and existing cache-fusion methods. Section~\ref{sec:discussion} concludes the paper.

\section{Preliminaries and Related Work}
\label{sec:prelim}
This section defines the RAG KV cache fusion problem and reviews token-selection methods for selective recomputation.
Table~\ref{tab:cache_fusion_notation} summarizes the notation, and Table~\ref{tab:selection_strategy_comparison} compares existing selection strategies.

\begin{table}[t]
    \caption{Notation for RAG KV cache fusion.}
    \small
    \centering
    \color{black}
    \setlength{\tabcolsep}{4pt}
    \renewcommand{\arraystretch}{1.08}
    \begin{tabular}{p{0.22\linewidth}|p{0.70\linewidth}}
        \hline
        \textbf{Notation} & \textbf{Definition} \\ \hline
        $U$ & User query appended after $C$ at serving time. \\ \hline
        $c_i$ & The $i$-th retrieved-evidence chunk in the request context. \\ \hline
        $C$ & Retrieved context $[c_1;c_2;\ldots;c_m]$ for the request, excluding $U$. \\ \hline
        $m$ & Number of retrieved chunks in one request. \\ \hline
        $N$ & Total number of context tokens in $C$, $N=\sum_{i=1}^{m}|c_i|$. \\ \hline
        $L$ & Number of Transformer layers in the LLM. \\ \hline
        $l$ & Layer index. \\ \hline
        $KV^{Full}(C)$ & Full-prefill KV cache of the retrieved context $C$, used as the quality reference. \\ \hline
        $\Pi_i(\cdot)$ & Position-remapping operation for placing the key cache of chunk $c_i$ at request positions while stitching its value cache in the same token order. \\ \hline
        $KV^{PIC}(C)$ & Request-level reused KV cache assembled from context chunk KV caches. \\ \hline
        $\mathcal{P}$ & Context token positions selected for recomputation. \\ \hline
        $KV^{New}_{p,l}(C)$ & Recomputed K/V entry for a selected position $p\in\mathcal{P}$ at layer $l$; it replaces the corresponding PIC entry in cache fusion. \\ \hline
        $KV^{Fuse}(C)$ & KV cache after selective recomputation. \\ \hline
        $\rho$ & Recomputation ratio, $\rho=|\mathcal{P}|/N$. \\ \hline
    \end{tabular}
    \label{tab:cache_fusion_notation}
\end{table}

\subsection{RAG KV Cache Fusion Problem}
\label{sec:pre_problem}

\subsubsection{RAG Context Construction}
\label{sec:pre_rag_reuse}
\label{sec:pre_full_prefill}

In data-management RAG services, each request combines retrieved evidence chunks from a relatively stable source corpus~\cite{zhao2024chat2data,chen2024singlestorev,balaka2025pneuma,deng2025alayadb}.
Across requests, the user query, retrieval order, and neighboring chunks may change, although the underlying retrieved chunks are reusable~\cite{agarwal2025cache,jin2024ragcache}.
For a user query $U$, we model the retrieved context as
\begin{equation}
    C = [c_1;\ c_2;\ \cdots;\ c_m],
    \label{eq:rag_prompt}
\end{equation}
where $c_1,\ldots,c_m$ are the ordered retrieved-evidence chunks~\cite{lewis2020retrieval,gao2023retrieval,guu2020retrieval,karpukhin2020dense}, and $U$ is appended after $C$ at serving time.
Thus, $C$ is the retrieved context excluding $U$, with $N=\sum_{i=1}^{m}|c_i|$ tokens.
Cache fusion is defined over the KV cache of $C$. Reused and recomputed K/V entries are indexed by context token positions.
The user query $U$ is processed online as the suffix that drives generation; it is not part of the reusable context cache that cache fusion assembles or refreshes.
Across all methods, the user query $U$ is processed online after $C$; the comparison focuses on how the retrieved-context cache for $C$ is materialized or refreshed.

Full prefill over $C$ materializes the layer-wise context KV cache~\cite{vaswani2017attention,dao2022flashattention,kwon2023pagedattention}.
We denote the full-prefill cache as
\begin{equation}
    KV^{Full}(C)=\{(\mathbf{K}^{l}_{C},\mathbf{V}^{l}_{C})\}_{l=1}^{L}
    \label{eq:full_prefill_cache}
\end{equation}
We use this cache as the full-prefill reference because all context chunks are contextualized together under the current request order and positions.
Because materializing this cache dominates long-context prefill time, optimizing prefill and KV caches has become a shared focus in recent LLM serving systems~\cite{kwon2023pagedattention,agrawal2024taming,zhong2024distserve,jiang2024minference}.

\subsubsection{Direct PIC Reuse}
\label{sec:pre_pic}

RAG KV cache fusion reduces serving-time prefill cost by precomputing chunk KV caches before serving.
Before serving, the system independently precomputes and stores the position-independent cache (PIC) of each reusable corpus chunk $c$~\cite{gim2024prompt,liu2024cachegen,liu2025lmcache}:
\begin{equation}
    KV_c^{PIC} \leftarrow \operatorname{LLM}(c),
    \label{eq:pic_offline}
\end{equation}
When a stored chunk $c$ is retrieved as $c_i$ in the current request, we write its stored cache as $KV_i^{PIC}$.
Each $KV_i^{PIC}$ is built independently, so its key cache is encoded at chunk-local positions starting from zero.
At serving time, the runtime retrieves the chunk KV caches needed by the current request, remaps them to their request positions, and stitches them according to the context order in $C$:
\begin{equation}
    KV^{PIC}(C)
    =
    [\Pi_1(KV_1^{PIC});\Pi_2(KV_2^{PIC});\ldots;\Pi_m(KV_m^{PIC})],
    \label{eq:pic_online}
\end{equation}
where $\Pi_i(\cdot)$ denotes position remapping for rotary or relative position embeddings~\cite{shaw2018self,su2023roformerenhancedtransformerrotary,press2022train}: the key cache is remapped from chunk-local positions to request positions, while the value cache is stitched in the same token order.
Direct reuse avoids recomputing chunk tokens, but it does not match full prefill:
\begin{equation}
    KV^{PIC}(C) \neq KV^{Full}(C)
    \label{eq:pic_not_equiv}
\end{equation}
This mismatch remains because position remapping fixes placement, not context.
Each reused $KV_i^{PIC}$ is built for one chunk in isolation, so it does not encode attention to the chunks that precede $c_i$ in the current request.
Full prefill, by contrast, lets each token attend to earlier chunks under the current context order and request positions.

\subsubsection{Selective KV Recomputation}
\label{sec:pre_blending}
\label{sec:pre_pipeline}

Selective recomputation mitigates the mismatch of direct PIC reuse without paying the full cost of full prefill.
Instead of recomputing every context token K/V entry, it selects context token positions $\mathcal{P}$ for recomputation and refreshes only the selected K/V entries under the current context order and positions~\cite{yao2025cacheblend}.
Let
\begin{equation}
    \rho = \frac{|\mathcal{P}|}{N},
    \label{eq:recompute_ratio}
\end{equation}
be the recomputation ratio.
For a selected position $p$ from $\mathcal{P}$, layer-$l$ recomputation produces $KV^{New}_{p,l}(C)$ under the current request order and positions; this entry is later written into the fused cache.
This per-token recomputation still follows causal attention: the selected token reads prefix positions $j\leq p$ through a partially fused prefix cache, where unselected prefix positions provide their reused $KV^{PIC}$ entries and already refreshed prefix positions in $\mathcal{P}$ provide their $KV^{New}$ entries.
The fused cache replaces only the selected PIC entries:
\begin{equation}
    KV^{Fuse}_{p,l}(C)
    =
    \begin{cases}
        KV^{New}_{p,l}(C), & p\in\mathcal{P},\\
        KV^{PIC}_{p,l}(C), & p\notin\mathcal{P}
    \end{cases}
    \label{eq:fuse_cache}
\end{equation}
\begin{figure}[t]
    \centering
    \includegraphics[width=\linewidth]{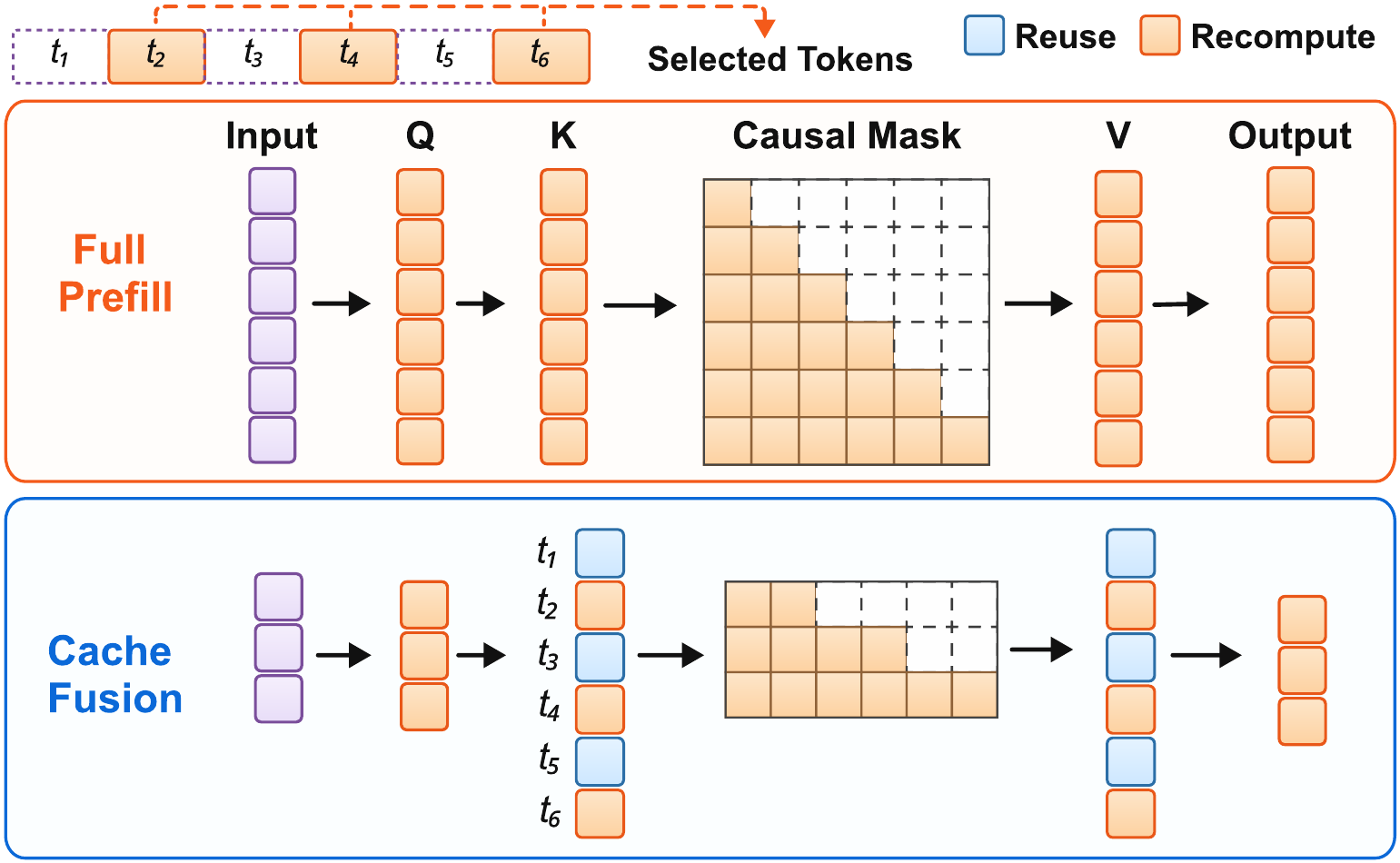}
    \caption{Full prefill computes all token states, whereas selective recomputation updates only selected K/V entries while reusing the remaining cache.}
    \Description{One-layer comparison of full prefill, direct position-independent cache reuse, and selective recomputation. Full prefill derives Q, K, and V for all context tokens under the current context order and positions. Direct reuse keeps precomputed chunk K and V states. Cache fusion recomputes Q, K, and V for selected tokens and merges the recomputed K and V entries with reused PIC entries to form the fused KV cache.}
    \label{fig:pre_recomp}
\end{figure}
Figure~\ref{fig:pre_recomp} compares full prefill and cache fusion at one Transformer layer.
Full prefill computes Q/K/V projections and attention outputs for every context token, thereby materializing the corresponding layer entries of $KV^{Full}(C)$.
Direct reuse performs no online recomputation and keeps all stitched PIC entries unchanged.
Cache fusion lies between these endpoints: the selected positions, e.g., $t_2$, $t_4$, and $t_6$, are recomputed under the current context order and positions, while unselected positions reuse their PIC entries.
The recomputed K/V entries are written back to the same context token positions, so the token order is unchanged and only selected entries differ from $KV^{PIC}(C)$.
The resulting cache is $KV^{Fuse}(C)$, and $\rho$ controls how many token positions are refreshed.

\subsection{Related Work on Token Selection}
\label{sec:pre_token_selection}

Existing cache-fusion systems differ mainly in how they choose $\mathcal{P}$.
We group them by query awareness and pipeline readiness; related long-context inference work also studies token/KV selection~\cite{zhang2023h2o,li2024snapkv,tang2024quest,xiao2024infllm}.

\subsubsection{Query-Agnostic Selection}

Query-agnostic selectors do not condition the mask on the current user query.
CacheBlend~\cite{yao2025cacheblend} selects tokens with high KV cache deviation: during selective recomputation, it recomputes a broad candidate set in early layers, compares the recomputed KV cache with the reused KV cache, and gradually keeps the tokens with the largest KV cache deviation.
EPIC~\cite{hu2024epic} further reduces runtime selection cost with a static LegoLink rule.
This rule is based on the observation that tokens near chunk boundaries, especially the first few tokens of a stitched chunk, can behave like attention sinks after position-independent linking; EPIC therefore recomputes a fixed number of boundary tokens for each chunk.

\subsubsection{Final-Layer Query-Aware Selection}

Final-layer query-aware selection uses the user query but restricts the serving-time evidence view to a single final-layer signal.
FusionRAG~\cite{Wang2026FromPC} first runs a query-only forward probe to obtain final-layer user-query representations.
It then scores context tokens by applying this final-layer query signal to the precomputed context keys and selects the top-ranked tokens for recomputation.
This avoids a full-context, multi-layer probing pass and keeps selection lightweight.
We focus on this token-selection rule; FusionRAG's other system optimizations are orthogonal to the selection-strategy comparison.
However, because the query signal is obtained without conditioning on the retrieved context and is taken only from the final layer, it may miss multi-hop evidence~\cite{trivedi2022musiquemultihopquestionssinglehop,yang2018hotpotqadatasetdiverseexplainable} and can be biased by non-evidence or attention-sink effects~\cite{jain2019attention,xiao2024streamingllm}.

\subsubsection{Full-View Query-Aware Selection}

Full-view query-aware selection uses full-context, all-layer signals.
ProphetKV~\cite{Wang2026ProphetKVUS} treats the user query as a predictor of which context tokens decoding will later use.
It lets user-query tokens attend over the retrieved context, aggregates user-query-to-context attention across user-query tokens and layers, and selects the top-scoring context tokens for recomputation.
This gives a relevance estimate that is closer to the current request than static or final-layer-only selectors, but the selection depends on broad context and layer visibility before recomputation starts, so selection can become a blocking stage in a layer-wise cache-fusion pipeline.

\newcommand{\preYes}{\ding{51}}
\newcommand{\prePart}{\ensuremath{\triangle}}
\newcommand{\preNo}{\textemdash}
\newcommand{\preUCell}[1]{\makebox[0.045\linewidth][c]{#1}}
\newcommand{\prePipeCell}[1]{\makebox[0.075\linewidth][c]{#1}}
\begin{table}[t]
    \caption{Token-selection strategies for cache fusion. U denotes user-query evidence; Pipe denotes pipeline compatibility; $\triangle$ denotes partial support.}
    \small
    \centering
    \color{black}
    \setlength{\tabcolsep}{3.5pt}
    \renewcommand{\arraystretch}{1.15}
    \begin{tabular*}{\linewidth}{@{\extracolsep{\fill}}p{0.17\linewidth}|p{0.57\linewidth}|c|c@{}}
        \hline
        \textbf{Strategy} & \textbf{Selection Signal} & \preUCell{\textbf{U}} & \prePipeCell{\textbf{Pipe}} \\ \hline
        Direct reuse
        & No recomputation
        & \preUCell{\preNo} & \prePipeCell{\preYes} \\ \hline
        CacheBlend
        & Shallow-layer KV cache deviation
        & \preUCell{\preNo} & \prePipeCell{\preYes} \\ \hline
        EPIC
        & Static chunk-boundary tokens
        & \preUCell{\preNo} & \prePipeCell{\preYes} \\ \hline
        FusionRAG
        & Final-layer user-query-to-context attention
        & \preUCell{\prePart} & \prePipeCell{\preYes} \\ \hline
        ProphetKV
        & All-layer user-query-to-context attention
        & \preUCell{\preYes} & \prePipeCell{\preNo} \\ \hline
    \end{tabular*}
    \label{tab:selection_strategy_comparison}
\end{table}

Table~\ref{tab:selection_strategy_comparison} summarizes the resulting design space.
Effective token selection must satisfy two coupled requirements: the selected tokens should reflect the current user query, and the selected set must be available before layer-wise cache fusion consumes it.
Existing selectors satisfy these requirements only partially.
Query-agnostic and final-layer-only methods remain easy to pipeline, but their limited evidence view can allocate recomputation to tokens with weak request-specific utility.
Full-view query-aware methods provide stronger relevance signals, but they obtain those signals by exposing broad token and layer views before recomputation, turning selection into a serialized pre-fusion stage.
The next section therefore builds a pipeline-aware query-dependent selector that compresses both views before recomputation starts~\cite{jiang2023llmlinguacompressingpromptsaccelerated,jiang2024longllmlingua,xu2023recomp}.

\section{\pname{} System Design}
\label{sec:system_design}
\label{sec:method_problem}

\pname{} is a compressed-view selector for RAG KV cache fusion.
Its role is to form $\mathcal{P}$ early enough for layer-wise cache fusion to overlap cache loading with selective recomputation.
\pname{} keeps query-aware token selection, but uses two compact evidence views in place of full-context, all-layer analysis: chunk-anchor query probing for token-view conditioning and critical-layer localization for layer-view scoring.
Section~\ref{sec:framework_overview} first gives the end-to-end workflow.
Section~\ref{sec:query_selection} then formalizes query-aware token selection and explains the full-view serving bottleneck.
Sections~\ref{sec:anchor_context} and~\ref{sec:critical_layer} detail chunk-anchor query probing and critical-layer token localization.
Finally, Section~\ref{sec:qcfuse_cache_pipeline} integrates the resulting $\mathcal{P}$ into the layer-wise cache-fusion pipeline.

\subsection{Framework Overview}
\label{sec:framework_overview}

\begin{figure*}[t]
    \centering
    \includegraphics[width=\textwidth]{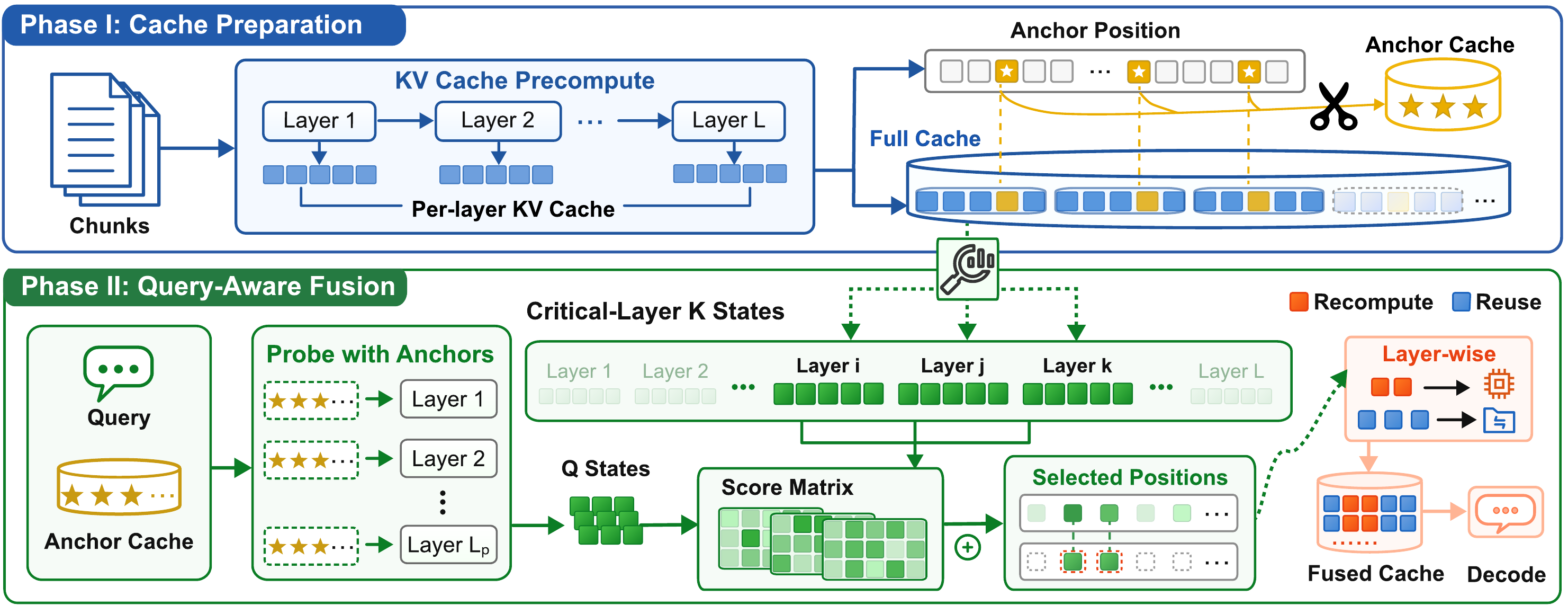}
    \caption{\pname{} forms query-aware recomputation masks using compact anchor tokens and a few critical layers.}
    \Description{System architecture diagram of QCFuse. The workflow shows PIC chunk KV cache construction, chunk-anchor KV cache construction, anchor-conditioned query probing, critical-layer key prefetching, selected-position recomputation, and final decoding from the fused KV cache.}
    \label{fig:Method}
\end{figure*}

Figure~\ref{fig:Method} organizes \pname{} into two phases.
The compressed-view selector is used only to form $\mathcal{P}$; cache fusion and decoding then proceed over the fused cache.
\begin{itemize}
    \item \textbf{Phase I: Cache Preparation.} \pname{} precomputes per-layer PIC KV caches for reusable chunks, selects anchor positions inside each chunk, profiles model-specific critical layers, and stores the reusable PIC cache, compact anchor cache, and critical-layer set.
    \item \textbf{Phase II: Query-Aware Fusion.} At serving time, \pname{} probes the query over the anchor cache, scores original context positions with critical-layer K states, and passes $\mathcal{P}$ to layer-wise cache fusion for sparse recomputation and decoding.
\end{itemize}

\subsection{Pipeline-Constrained Token Selection}
\label{sec:query_selection}
\label{sec:method_basic_idea}

\begin{figure}[!t]
    \centering
    \includegraphics[width=\linewidth]{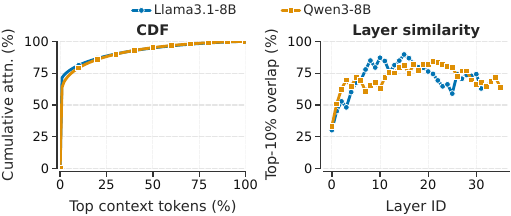}
    \caption{Full-view profiling reveals compression opportunities in both selector views: attention mass concentrates on a small token subset, and middle layers best approximate all-layer selection.}
    \Description{Two-panel profile on MuSiQue for Qwen3-8B and Llama3.1-8B. The left panel plots Cumulative attention score, the percentage of query-to-context attention mass covered by top-ranked context tokens. The right panel plots top-10 percent selected-token overlap between each layer and all-layer selection across layer IDs.}
    \label{fig:mq_cdf_similarity}
\end{figure}

\begin{figure*}[t]
    \centering
    \includegraphics[width=\textwidth]{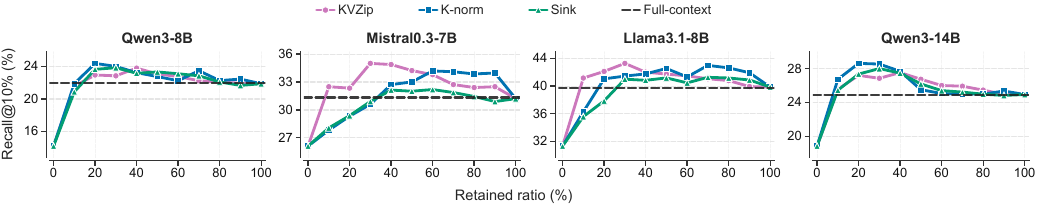}
    \caption{KVzip@10\% provides a compact operating point near the full-context reference while using only a small anchor cache.}
    \Description{Chunk-anchor profiling plot for QCFuse. Each panel corresponds to one model. The x-axis sweeps retained token ratio and the y-axis reports Recall at 10 percent. Curves compare Sink, Knorm, and KVzip anchor ranking rules, and the dashed line denotes the full-context probing reference.}
    \label{fig:anchor_profile}
\end{figure*}

This subsection defines the selection objective used by \pname{}.
We instantiate a ProphetKV-style full-view selector as an expensive attention-based reference~\cite{Wang2026ProphetKVUS}, then use it to expose where token and layer views can be compressed.
The goal is not to solve an online optimization problem, but to define the requirements that the offline anchor and layer profiles must satisfy.

\pname{} frames the selection as a pipeline-constrained approximation of full-view evidence coverage.
Given request query $U$ and retrieved context $C$, the target is to choose a recomputation set $\mathcal{P}$ that covers query-relevant evidence while exposing the selector to compact token and layer views.
We denote these views as $C_{\mathcal{A}}$ and $\mathcal{L}^{\star}$; the full-view selector uses $C$ and the full layer set $\mathcal{L}=\{1,\ldots,L\}$.
The following formulation is a design objective: offline preparation fixes the per-chunk anchors that instantiate $C_{\mathcal{A}}$, offline profiling fixes $\mathcal{L}^{\star}$, and online selection only assembles $C_{\mathcal{A}}$ for the retrieved chunks and estimates $\mathcal{P}$.
\begin{equation}
\begin{aligned}
\max_{\mathcal{P},\,C_{\mathcal{A}},\,\mathcal{L}^{\star}}
\quad & S(\mathcal{P};U,C) \\
\text{s.t.}\quad
& |\mathcal{P}|=\lfloor\rho N\rfloor,\quad
|C_{\mathcal{A}}|\leq r_a N,\quad
|\mathcal{L}^{\star}|=k_l, \\
& T_{\mathrm{select}}(C_{\mathcal{A}},\mathcal{L}^{\star})
< T_{\mathrm{select}}(C,\mathcal{L}).
\end{aligned}
\label{eq:pipeline_constrained_selection}
\end{equation}
Here $S(\mathcal{P};U,C)$ denotes an offline evidence-coverage criterion: it measures how much query-relevant evidence in the original context $C$ is covered by the selected positions $\mathcal{P}$.
This criterion is not observable or optimized online; it defines what the offline profiles should approximate.
$C_{\mathcal{A}}$ and $\mathcal{L}^{\star}$ only constrain what the online selector can inspect when estimating $\mathcal{P}$.
The constraints fix the recomputation budget, bound the token and layer views, and require selection with $(C_{\mathcal{A}},\mathcal{L}^{\star})$ to be faster than selection with the full context $C$ and all layers $\mathcal{L}$.
A ProphetKV-style full-view selector supplies the expensive attention-based reference signal for this profiling criterion.
Span-labeled calibration then checks whether compact evidence views preserve evidence coverage under the same selection budget.

\subsubsection{Full-View Query-Aware Selection}
\label{sec:query_selection_full_view}
A ProphetKV-style full-view selector forwards the user query $U$ with the complete stitched $KV^{PIC}(C)$ from Eq.~\ref{eq:pic_online} as prefix.
For each layer $l$, this produces context-conditioned query states:
\begin{equation}
    \mathbf{Q}_{U}^{l}(C)
    \leftarrow
    \operatorname{LLM}^{l}\left(U;KV^{PIC}(C)\right)
    \label{eq:selector_query_state}
\end{equation}
This probe reads reused context entries only for selection; it does not refresh any context K/V entry.

The selector then scores original context positions by comparing $\mathbf{Q}_{U}^{l}(C)$ with the keys $\mathbf{K}_{C}^{PIC,l}$.
At layer $l$, $\operatorname{Attn}^{l}\!\left(\mathbf{Q}_{U}^{l}(C),\mathbf{K}_{C}^{PIC,l}\right)_t$ denotes the user-query-to-context attention mass assigned to context token position $t$ in $C$, aggregated over user-query tokens and heads.
The full-view importance score sums this signal over all layers:
\begin{equation}
    I(t)
    =
    \sum_{l=1}^{L}
    \operatorname{Attn}^{l}
    \left(\mathbf{Q}_{U}^{l}(C),\mathbf{K}_{C}^{PIC,l}\right)_t
    \label{eq:importance}
\end{equation}
Given recomputation ratio $\rho$, the selector returns $\mathcal{P}(U,C)$, the top $\lfloor\rho N\rfloor$ context token positions ranked by $I(t)$:
\begin{equation}
    \mathcal{P}(U,C)
    =
    \operatorname{TopK}
    \left(
        I(t),\ \lfloor \rho N \rfloor
    \right)
    \label{eq:critical_mask}
\end{equation}

\subsubsection{Full-View Bottleneck and Compression Opportunity}
\label{sec:query_selection_bottleneck}

The reference is useful because it has both missing views: query states conditioned on the complete reused context and all-layer attention signals for localization.
It is also the wrong shape for the serving critical path.
Sparse recomputation cannot start until $\mathcal{P}$ is known, and computing $\mathcal{P}$ requires full-context key signals from all layers.
When layer-wise context keys reside in CPU memory or SSD, selection transfers them to the GPU as serialized pre-fusion work rather than letting the cache-fusion pipeline overlap loading with recomputation.
This bottleneck motivates the latency constraint in Eq.~\ref{eq:pipeline_constrained_selection}.

The compression opportunity is that selector evidence is not uniformly useful across tokens or layers.
Long-context studies and sparse-attention methods suggest that only a subset of context tokens usually receives substantial attention mass, while many tokens contribute little to the next representation~\cite{liu2024lost,yuan2025native,liu2023scissorhands,yang2024pyramidinfer}.
Model probing studies further show that Transformer layers encode different linguistic and semantic signals, so a token-localization decision may not require evidence from every layer~\cite{tenney2019bert,jawahar2019does,vig2019analyzing,skean2025layer}.
We profile the ProphetKV-style full-view selector on MuSiQue for Qwen3-8B and Llama3.1-8B to check whether the same pattern appears in our selection setting.
Figure~\ref{fig:mq_cdf_similarity} reports token-view concentration in the left panel and single-layer agreement with all-layer selection in the right panel.
For the token-view diagnostic, following the cumulative-attention-score view used by H2O~\cite{zhang2023h2o,cai2024pyramidkv}, we rank context token positions by the all-layer importance score $I(t)$ from Eq.~\ref{eq:importance} and plot the \emph{Cumulative attention score} (\%), i.e., the percentage of query-to-context attention mass covered by the top-ranked tokens.
For the layer-view diagnostic, we inspect the unaggregated layer-wise score $\operatorname{Attn}^{l}$ before the all-layer sum.
For each layer, we select the top-10\% context token positions by $\operatorname{Attn}^{l}$ and compare them with the top-10\% all-layer reference set selected by $I(t)$.

Figure~\ref{fig:mq_cdf_similarity} exposes two empirical patterns: cumulative attention rises steeply for top-ranked tokens, and all-layer agreement is strongest in middle layers.

\noindent\textbf{Implication 1.} \emph{The selector view can be compressed along both dimensions: top-ranked tokens carry concentrated attention mass, and middle layers best approximate all-layer localization.}

\begin{figure*}[t]
    \centering
    \includegraphics[width=\textwidth]{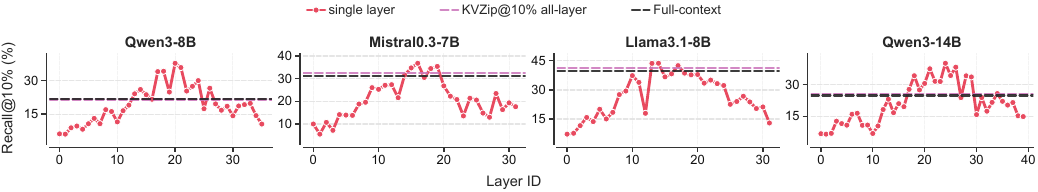}
    \caption{Single-layer evidence localization peaks in model-dependent middle layers rather than final layers.}
    \Description{Single-layer localization profiling plot for QCFuse. The plot sweeps layer IDs across models and compares Recall at 10 percent with all-layer and full-context references.}
    \label{fig:layer_profile}
\end{figure*}

\begin{figure}[t]
    \centering
    \includegraphics[width=\linewidth]{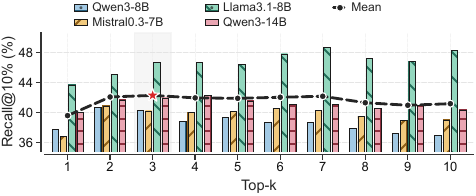}
    \caption{Top-3 profiled layers capture most of the Recall@10\% gain; \pname{} uses them as the serving-time layer view.}
    \Description{Few-layer fusion profiling plot for QCFuse. The plot compares evidence-localization quality as the number of selected top-ranked layers increases.}
    \label{fig:topk_profile}
\end{figure}

\subsubsection{Evidence-Guided Compression Calibration}
\label{sec:query_selection_span_profile}
The compression profile shows that smaller evidence views may be sufficient, but \emph{attention mass is still an estimated signal rather than a gold evidence label}.
\pname{} therefore calibrates its compressed token and layer views with evidence localization: a useful compressed view should keep answer-bearing evidence inside the fixed selection budget.
For this calibration, \pname{} uses SQuAD~\cite{rajpurkar-etal-2016-squad}, NewsQA~\cite{trischler-etal-2017-newsqa}, and Natural Questions~\cite{kwiatkowski-etal-2019-natural}, whose samples include explicit answer spans.
We use these span-labeled datasets because they provide explicit answer-position spans for profiling; this makes the profile a model-level selector diagnostic rather than tuning on the end-to-end evaluation benchmarks in Section~\ref{sec:exp}.
For each model, this offline calibration fixes the default anchor view and critical-layer budget before end-to-end QA evaluation, and Section~\ref{sec:exp_ablation} later validates these defaults through ablation.
We map each annotated answer span to context token positions and merge all spans in the same sample into a gold answer-position set $E$.
For each profile request, a selector variant assigns an importance score $I(t)$ to every context token position $t$.
For the profiling metric only, we use a 10\% selection budget and instantiate the selector output as the highest-scoring context token positions,
\begin{equation}
    \mathcal{P}_{10\%}
    =
    \operatorname{TopK}
    \left(
        I(t), \lfloor 10\% \cdot N \rfloor
    \right)
    \label{eq:profile_top10}
\end{equation}
We use 10\% as this offline profiling selection budget because the full-view profile shows a steep attention-mass concentration at small token ratios.
This profiling budget is independent of the serving recomputation ratio $\rho$, which is varied in the end-to-end evaluation.

Given $\mathcal{P}_{10\%}$ and the gold answer-position set $E$, the profile reports Recall@10\% as the fraction of answer positions covered by the selected top-10\% context token positions:
\begin{equation}
    \operatorname{Recall@10\%}
    =
    \frac{|\mathcal{P}_{10\%}\cap E|}{|E|}
    \label{eq:profile_recall_at10}
\end{equation}
Recall@10\% matches the selector's objective: the selected set need not contain only answer tokens, but it should cover answer-bearing evidence positions under a fixed token budget.
This calibration provides a shared selection-quality criterion for both compression decisions: Section~\ref{sec:anchor_context} uses it to choose the anchor ranking rule and retained ratio, and Section~\ref{sec:critical_layer} uses it to choose the critical-layer set.
End-to-end QA experiments later evaluate final quality and latency.

\subsection{Chunk-Anchor Query Probing}
\label{sec:anchor_context}

Chunk-anchor query probing instantiates the compressed token view $C_{\mathcal{A}}$ in Eq.~\ref{eq:pipeline_constrained_selection}.
Query-only probing misses retrieved evidence, while full-context probing recreates the selector-side loading cost.
Per-chunk anchors give every retrieved chunk a compact conditioning path while keeping the final recomputation set over original context positions.
The rest of this subsection is organized as follows: Section~\ref{sec:anchor_construction} constructs per-chunk anchors, Section~\ref{sec:anchor_probe} uses the anchor cache to condition the query while preserving original context positions, and Section~\ref{sec:anchor_profile_default} profiles anchor rules and ratios to choose the default token view.

\subsubsection{Per-Chunk Anchor Construction}
\label{sec:anchor_construction}
The anchor view should preserve enough reusable chunk evidence for query conditioning while keeping the probing cache small.
Following the context definition in Section~\ref{sec:pre_rag_reuse}, anchors are defined over reusable corpus chunks that may later be retrieved into $C$.
\pname{} constructs this view independently for each chunk, so a single long or high-scoring chunk cannot consume the entire probing budget of a request.

For each reusable chunk $c_i$, an anchor selection rule assigns a score $G_i(r)$ to each chunk-local token position $r$.
Let $\operatorname{TopK}(G_i,k)$ return the $k$ highest-scoring positions in $c_i$ under $G_i$.
Given retained ratio $r_a$, the anchor position set is
\begin{equation}
    \mathcal{A}_i
    =
    \operatorname{TopK}
    \left(
        G_i,\ \lceil r_a |c_i| \rceil
    \right)
    \label{eq:anchor_set}
\end{equation}
KV-cache compression benchmarks cover a broad range of pruning rules and provide a common reference for comparing practical KV selection methods~\cite{devoto2025expectedattention,ge2023model,feng2024adakv,zhang2025pqcache}.
We therefore instantiate three representative offline ranking rules for $G_i(r)$: a static sink-style rule, a lightweight key-statistics heuristic, and a strong reconstruction-based method.
\begin{itemize}
    \item \textbf{Sink}~\cite{xiao2024streamingllm,hu2024epic} is a lightweight boundary heuristic that keeps sink-style tokens near the beginning of each chunk.
    \item \textbf{Knorm}~\cite{devoto-etal-2024-simple} ranks token positions by the norm of their key-cache tensors.
    \item \textbf{KVzip}~\cite{kim2025kvzip} runs a self-supervised repeat-the-context reconstruction probe over the cached context KV and uses the attention received by each KV entry.
\end{itemize}
All three scores can be collected during offline cache preparation together with PIC KV construction.

\subsubsection{Anchor-Conditioned Query Probing}
\label{sec:anchor_probe}
During profiling and serving, \pname{} preserves the original order and request position IDs of retained anchors while exposing only anchor positions from each retrieved chunk to the query probe.
The resulting chunk-anchor context $C_{\mathcal{A}}$ is:
\begin{equation}
    C_{\mathcal{A}}
    =
    [c_1[\mathcal{A}_1];\ c_2[\mathcal{A}_2];\ldots;c_m[\mathcal{A}_m]]
    \label{eq:anchor_probe_sequence}
\end{equation}
The anchor cache is extracted from the corresponding PIC cache entries:
\begin{equation}
\begin{aligned}
    KV_i^{\mathcal{A}}
    &=
    KV_i^{PIC}[\mathcal{A}_i], \\
    KV^{\mathcal{A}}(C_{\mathcal{A}})
    &=
    [\Pi_1(KV_1^{\mathcal{A}});\ldots;\Pi_m(KV_m^{\mathcal{A}})].
\end{aligned}
    \label{eq:anchor_cache}
\end{equation}
Using the stitched anchor KV cache $KV^{\mathcal{A}}(C_{\mathcal{A}})$ as the prefix, the query probe computes anchor-conditioned query states:
\begin{equation}
    \mathbf{Q}_{U}^{l}(C_{\mathcal{A}})
    \leftarrow
    \operatorname{LLM}^{l}
    \left(U;KV^{\mathcal{A}}(C_{\mathcal{A}})\right)
    \label{eq:anchor_query_state}
\end{equation}
The anchor tokens are fed as a compact subsequence but keep their original request position IDs for rotary or relative position handling, and the user-query position IDs remain those after the full context $C$.
The chunk-anchor context only conditions the query states; the following scoring step still ranks token positions in the original context $C$ for recomputation.

\subsubsection{Anchor Profile and Default View}
\label{sec:anchor_profile_default}

The anchor profile fixes the chunk-local ranking rule and retained ratio $r_a$.
Figure~\ref{fig:anchor_profile} compares Sink, Knorm, and KVzip under the same all-layer scoring and Recall@10\% protocol, with full-context probing as the reference.
Recall@10\% increases quickly once a small anchor ratio is retained.
At the anchor retained ratio $r_a=0.1$, KVzip remains near the full-context reference while loading only a compact anchor cache for query probing.

\noindent\textbf{Implication 2.} \emph{KVzip@10\% ($r_a=0.1$) is the selected anchor view because it captures the early Recall@10\% gain while keeping the probing cache small.}

\subsection{Critical-Layer Token Localization}
\label{sec:critical_layer}

Critical-layer token localization instantiates the compressed layer view $\mathcal{L}^{\star}$ in Eq.~\ref{eq:pipeline_constrained_selection}.
The following profile shows that final-layer-only scoring is not the strongest localization signal, while all-layer scoring is informative but pipeline-blocking.
\pname{} therefore profiles a small critical-layer set that best localizes answer-bearing evidence while avoiding all-layer context-key loading.
The rest of this subsection is organized as follows: Section~\ref{sec:critical_layer_single_profile} profiles single-layer evidence localization, Section~\ref{sec:critical_layer_set} builds candidate critical-layer sets, and Section~\ref{sec:critical_layer_default_view} selects the serving-time layer view.

\subsubsection{Single-Layer Localization Profile}
\label{sec:critical_layer_single_profile}
For each model, \pname{} fixes the KVzip@10\% anchor view from Section~\ref{sec:anchor_context} so the offline profile isolates the layer-view effect.
For a layer $l$, the selector uses anchor-conditioned query states $\mathbf{Q}_{U}^{l}(C_{\mathcal{A}})$ and the PIC context keys at the original context positions, $\mathbf{K}_{C}^{PIC,l}$, to assign a single-layer score to each context token position $t$:
\begin{equation}
    I_l(t)
    =
    \operatorname{Attn}^{l}
    \left(\mathbf{Q}_{U}^{l}(C_{\mathcal{A}}),\mathbf{K}_{C}^{PIC,l}\right)_t
    \label{eq:single_layer_importance}
\end{equation}
Using $I_l(t)$, the profile selects the top-10\% context token positions and computes Recall@10\% with the evidence-guided calibration protocol from Section~\ref{sec:query_selection_span_profile}.
Let $R(l)$ denote this Recall@10\% score; it measures how well layer $l$ alone localizes answer-bearing evidence positions.
Because the anchor view and token budget are fixed, differences in $R(l)$ reflect layer-view localization strength rather than token-view configuration.

\subsubsection{Critical-Layer Set Construction}
\label{sec:critical_layer_set}
The single-layer profile answers which layers are individually useful, but it does not determine how much layer visibility is sufficient.
We therefore sweep a layer budget $k$ and form a candidate set from the $k$ layers with the highest localization scores:
\begin{equation}
    \mathcal{L}_{k}
    =
    \operatorname{TopK}
    \left(
        R(l),\ k
    \right)
    \label{eq:critical_layer_set}
\end{equation}
where $\operatorname{TopK}$ is taken over layer IDs $l\in\{1,\ldots,L\}$.

For any candidate set $\mathcal{L}_{k}$, the selector uses only the PIC context keys from those layers.
With the anchor-conditioned query states fixed, the token-importance score becomes:
\begin{equation}
    I_{\mathcal{L}_{k}}(t)
    =
    \sum_{l\in\mathcal{L}_{k}}
    \operatorname{Attn}^{l}
    \left(\mathbf{Q}_{U}^{l}(C_{\mathcal{A}}),\mathbf{K}_{C}^{PIC,l}\right)_t
    \label{eq:critical_layer_importance}
\end{equation}
This score instantiates $I(t)$ in Eq.~\ref{eq:critical_mask} to produce the selected recomputation set $\mathcal{P}$.

\subsubsection{Layer Profile and Default View}
\label{sec:critical_layer_default_view}
Figure~\ref{fig:layer_profile} reports the single-layer localization profile, and Figure~\ref{fig:topk_profile} reports the top-$k$ critical-layer aggregation profile.
The best single-layer localization appears in model-dependent middle layers.
The top-$k$ curve captures most of its Recall@10\% gain by $k=3$.
Thus, $k=3$ is the smallest profiled budget beyond which additional layers bring limited measured gain.
We set the serving-time critical-layer set to $\mathcal{L}^{\star}=\mathcal{L}_{3}$.

\noindent\textbf{Implication 3.} \emph{Top-3 profiled middle layers are the selected layer view because they keep most measured Recall@10\% gain while avoiding all-layer KV loading.}

\subsection{Pipeline-Integrated Cache Fusion}
\label{sec:qcfuse_cache_pipeline}

\begin{figure}[t]
    \centering
    \includegraphics[width=\linewidth]{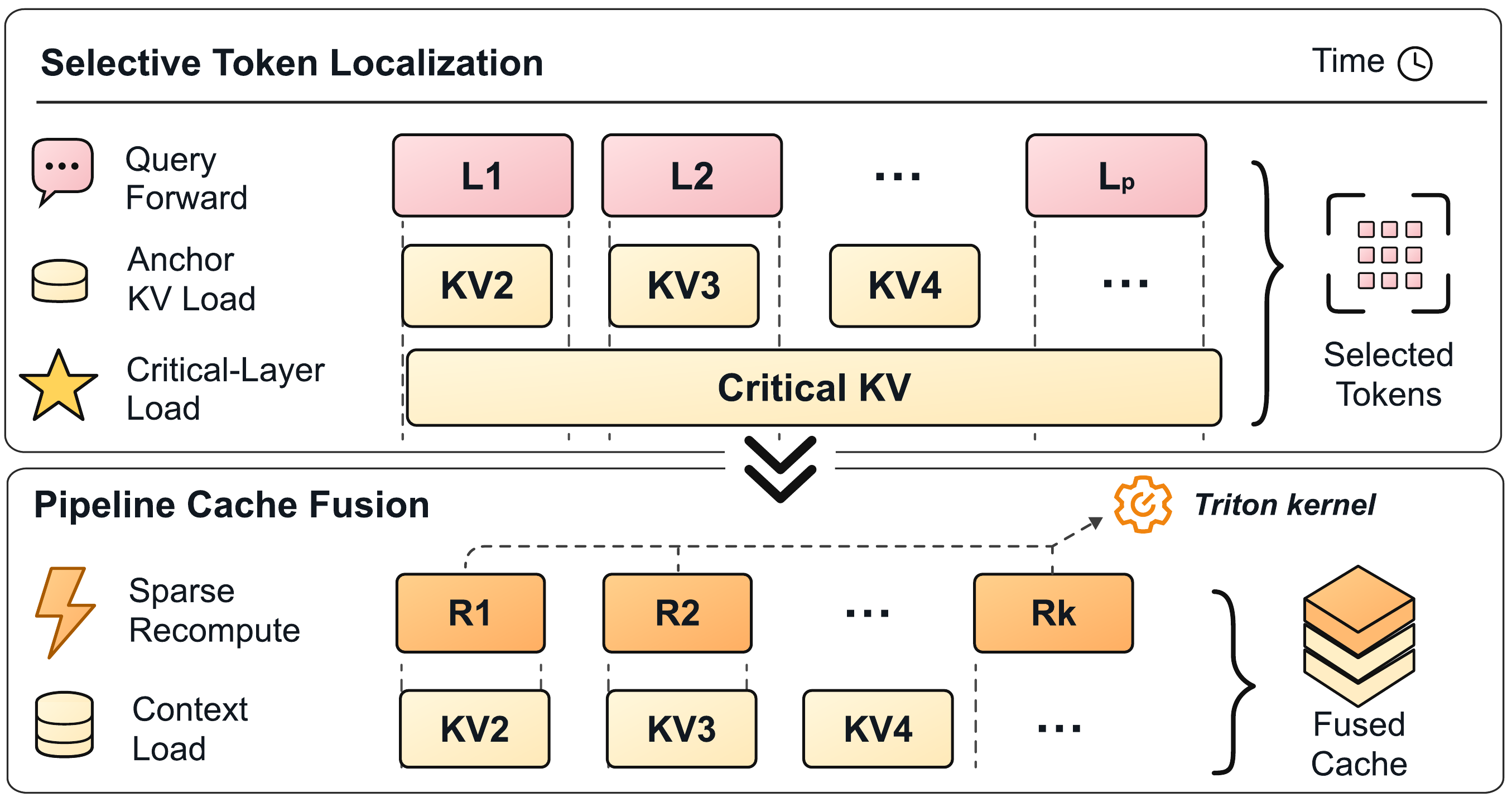}
    \caption{\pname{} shortens the pre-fusion selection path and then overlaps layer-wise sparse recomputation with KV-cache loading.}
    \Description{Runtime pipeline of QCFuse. The figure shows offline cache preparation, online chunk-anchor query probing, critical-layer key prefetching, selected-position set construction, and layer-wise selective recomputation overlapped with loading reused KV caches for later layers.}
    \label{fig:qcfuse_pipeline}
\end{figure}

\pname{} integrates the selected-position set $\mathcal{P}$ into the online cache-fusion pipeline in Figure~\ref{fig:qcfuse_pipeline}.

The compressed-view selector changes only how $\mathcal{P}$ becomes available; the cache-fusion semantics remain the same as selective recomputation in Section~\ref{sec:pre_blending}.
The runtime consumes the Phase-I artifacts: the PIC cache, the KVzip-selected 10\% anchor cache, and the profiled critical-layer set.
Compared with full-view selection, this reduces selection-time data movement: query probing loads only the per-chunk anchor cache, and token scoring needs full-context keys only for $\mathcal{L}^{\star}$ rather than for all $L$ layers.
The remaining PIC entries are loaded by the normal layer-wise fusion schedule.
For each online request, Figure~\ref{fig:qcfuse_pipeline} separates the runtime into two stages:
\begin{itemize}
    \item \textbf{Selective token localization.} The runtime loads anchor KV caches, stitches them into $C_{\mathcal{A}}$, and computes query states up to the deepest layer in $\mathcal{L}^{\star}$. In parallel, it prefetches PIC context keys only for layers in $\mathcal{L}^{\star}$. These signals score original context positions and produce $\mathcal{P}$.
    \item \textbf{Pipeline cache fusion.} Once $\mathcal{P}$ is ready, each layer starts from the loaded PIC entries. A Triton kernel recomputes only positions in $\mathcal{P}$ using the current fused prefix cache, scatters the refreshed K/V entries back to the same positions, and overlaps this sparse recomputation with KV-cache loading for later layers.
\end{itemize}
By avoiding full-context, all-layer selection, this compressed-view localization stage can run before Triton sparse recomputation while KV-cache loading overlaps with later layers.

\section{Experiments}
\label{sec:exp}

We evaluate \pname{} through five questions that follow the design claims in Sections~\ref{sec:prelim} and~\ref{sec:system_design}.

\begin{itemize}
    \item \textbf{End-to-End Performance.} How much does \pname{} reduce TTFT compared with full prefill and cache-fusion baselines~\cite{yao2025cacheblend,hu2024epic,Wang2026FromPC,Wang2026ProphetKVUS} while preserving task quality? (Sec.~\ref{sec:exp_e2e})
    \item \textbf{Long-Context Scalability.} Can \pname{} maintain its quality advantage as the retrieved context grows? (Sec.~\ref{sec:exp_long_context})
    \item \textbf{Bandwidth Sensitivity.} How does \pname{} behave when cache-loading bandwidth becomes the bottleneck, compared with full-context user-query-aware selection? (Sec.~\ref{sec:exp_bandwidth})
    \item \textbf{Serving Throughput.} Under increasing request load, how far can \pname{} sustain low TTFT compared with competing serving strategies? (Sec.~\ref{sec:exp_throughput})
    \item \textbf{Design Ablation.} How much do chunk-anchor query probing and critical-layer selection contribute across recomputation budgets? (Sec.~\ref{sec:exp_ablation})
\end{itemize}

\subsection{Experimental Configuration}
\label{sec:exp_setup}

\subsubsection{Implementation}
We implement \pname{} in SGLang 0.5.4~\cite{zheng2023efficiently}. The implementation adds offline chunk KV cache construction and an online path for user-query-aware token selection followed by layer-wise selective KV cache recomputation. All evaluated serving strategies run in the same SGLang-based BF16 serving stack~\cite{zheng2023efficiently,kwon2023pagedattention}. The server has two NVIDIA H20 GPUs, 128GB of DRAM, and a 1TB NVMe SSD with 10GB/s sequential read bandwidth. Unless noted otherwise, all cache-fusion strategies use the same retrieved chunks, hardware budget, metrics, and recomputation-ratio sweep.

\subsubsection{Models}
Following the model scope of prior cache-fusion baselines~\cite{Wang2026FromPC,Wang2026ProphetKVUS}, we evaluate open-weight decoder-only LLMs because cache fusion requires access to Transformer KV caches~\cite{vaswani2017attention,kwon2023pagedattention,shazeer2019fast,ainslie2023gqa}, cache placement, and the serving-time selection path. We use Mistral-v0.3-7B~\cite{jiang2023mistral7b}, Llama-3.1-8B~\cite{grattafiori2024llama}, and Qwen3-8B~\cite{qwen3technicalreport}. To test whether the same trends hold at a larger parameter scale, we also include Qwen3-14B~\cite{qwen3technicalreport}. These four models differ in architecture family and tokenizer, allowing us to test whether \pname{} generalizes beyond a single model family.

\subsubsection{Baselines}
Following prior cache-fusion evaluations~\cite{yao2025cacheblend,hu2024epic,Wang2026FromPC,Wang2026ProphetKVUS}, we include \textbf{Full prefill} and \textbf{Direct PIC reuse}~\cite{gim2024prompt} as the two basic references. Full prefill contextualizes all retrieved chunks together under the current request, so it provides the quality reference. Direct PIC reuse keeps independently computed chunk KV caches unchanged, so it provides the low-TTFT endpoint without selective recomputation.

We also include recent representative cache-fusion strategies, CacheBlend~\cite{yao2025cacheblend}, EPIC~\cite{hu2024epic}, FusionRAG~\cite{Wang2026FromPC}, and ProphetKV~\cite{Wang2026ProphetKVUS}. CacheBlend and EPIC represent user-query-agnostic recomputation rules that fit the layer-wise pipeline. FusionRAG uses a lightweight final-layer user-query-token signal. ProphetKV is the strongest user-query-aware baseline, but it builds its signal from broad context and layer visibility before recomputation. For a fair selection-strategy comparison, all cache-fusion methods use the same retrieved chunks, recomputation ratios, cache placement, and layer-wise recomputation pipeline; when a prior system includes additional optimizations, such as FusionRAG, we isolate its token-selection rule and do not attribute orthogonal system optimizations to the selector. Together, these baselines cover the key cache-fusion design choices, including whether the selector uses the user query, whether the selector is compatible with layer-wise pipelining, and how much KV cache must be loaded before recomputation starts.

\subsubsection{Datasets}
We evaluate two benchmark families following the evaluation setting used by prior cache-fusion baselines~\cite{Wang2026ProphetKVUS}, so that quality, TTFT, and throughput are measured under the same comparison protocol. LongBench~\cite{bai2024longbenchbilingualmultitaskbenchmark} tests semantic understanding over natural long-context inputs. We use MuSiQue~\cite{trivedi2022musiquemultihopquestionssinglehop}, 2WikiMQA~\cite{ho2020constructingmultihopqadataset}, and HotpotQA~\cite{yang2018hotpotqadatasetdiverseexplainable}, which require the model to combine evidence across retrieved chunks. RULER~\cite{hsieh2024ruler} provides controlled synthetic long-context tasks; we use multi-query retrieval (\texttt{mq}), multi-value retrieval (\texttt{mv}), and variable tracking (\texttt{vt}) to stress exact retrieval and state tracking. Together, these benchmarks cover both natural semantic QA and controlled retrieval cases commonly used to evaluate long-context RAG behavior. Following the standard RAG request construction and the baseline setup, we split each context into 512-token chunks and preserve the retrieved order when assembling the request. Each dataset contributes 200 samples. The main tradeoff uses 20 chunks per request, which preserves the available context for the evaluated samples while keeping all methods in the long-context cache-fusion regime.

\subsubsection{Metrics and recomputation ratios}
Following the cache-fusion baselines~\cite{yao2025cacheblend,hu2024epic,Wang2026FromPC,Wang2026ProphetKVUS}, we report task quality, TTFT, and serving throughput because long-context RAG serving must preserve answer correctness, first-token responsiveness, and capacity under concurrent request load. Quality is measured with the official F1 score for LongBench~\cite{bai2024longbenchbilingualmultitaskbenchmark} and string match (SM) for RULER~\cite{hsieh2024ruler}. We normalize both metrics across tasks and models, yielding Normalized-F1 and Normalized-SM in the tradeoff figures so that results from different benchmarks and model families can be compared on the same scale. TTFT measures the time before the first generated token and therefore captures the prefill-side cost targeted by cache fusion~\cite{kwon2023pagedattention,zhong2024distserve,agrawal2024taming}. It includes online token selection, required cache loading, selective recomputation, and first-token generation; offline cache preparation and profiling are excluded. Throughput is measured as average requests per second while tracking TTFT as request load increases~\cite{yu2022orca,zhong2024distserve,agrawal2024taming,lin2024qserve}. We denote the recomputation ratio by $\rho$, i.e., the fraction of retrieved-context token positions whose KV entries are recomputed under the current request. The main quality--TTFT tradeoff sweeps $\rho$ from $0.1$ to $0.5$ in increments of $0.1$; direct reuse is the $\rho=0$ endpoint, and full prefill is the $\rho=1$ reference.

\subsection{End-to-End Quality--TTFT Tradeoff}
\label{sec:exp_e2e}

\begin{figure*}[t]
  \centering
  \includegraphics[width=\linewidth]{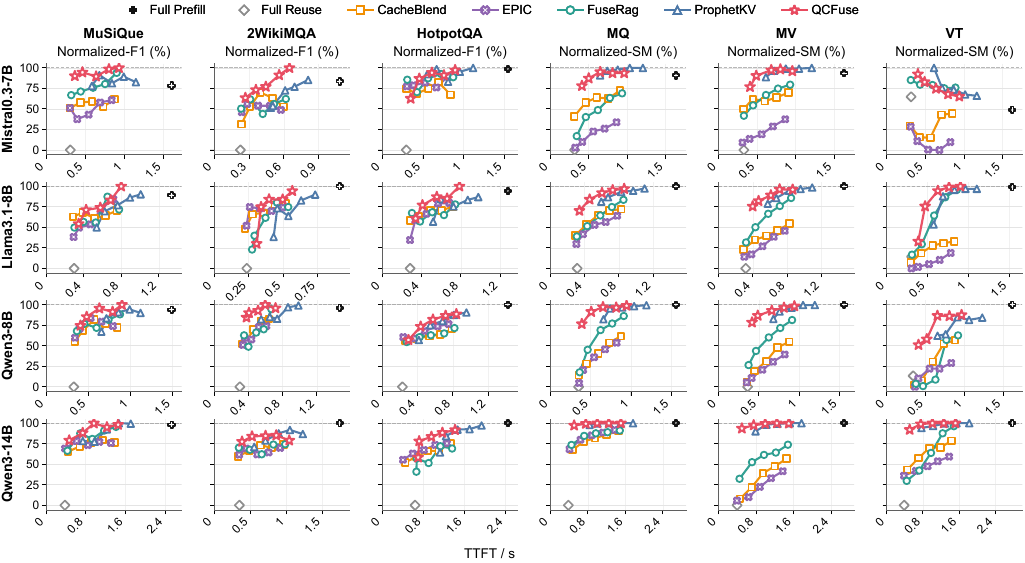}
  \caption{\pname{} reaches full-prefill-level quality at lower TTFT. Each panel shows one model--task pair.}
  \Description{A grid of quality versus time-to-first-token plots. Rows are evaluated models and columns are MuSiQue, 2WikiMQA, HotpotQA, mq, mv, and vt. Curves compare full prefill, direct reuse, QCFuse, ProphetKV, FusionRAG, CacheBlend, and EPIC across recomputation ratios.}
  \label{fig:final_exp}
\end{figure*}

This experiment tests whether \pname{} can recover full-prefill-level quality while reducing TTFT. Figure~\ref{fig:final_exp} shows the fine-grained quality--TTFT tradeoff across models, tasks, baselines, and recomputation ratios; Figure~\ref{fig:benchmark_aggregate_ttft_quality} summarizes the same evidence for LongBench and RULER. Points closer to the upper-left corner are better.

\subsubsection{Quality Recovery}
This comparison asks whether \pname{} selects the right tokens to recompute. LongBench requires connecting the query to evidence distributed across chunks, while RULER-style tasks require localizing exact keys, values, or variable states. As shown in Figure~\ref{fig:final_exp}, \pname{} moves into the high-quality region across both benchmark families. The aggregate view in Figure~\ref{fig:benchmark_aggregate_ttft_quality} shows that \pname{} is close to ProphetKV, above CacheBlend, EPIC, and FusionRAG, and in several semantic QA cases reaches or slightly exceeds full prefill. \textbf{\pname{} recovers full-prefill-level quality because its recomputation budget is guided by the user query rather than by request-independent cache signals.}

\subsubsection{TTFT Reduction}
This comparison also asks whether query-aware selection delays serving. Figure~\ref{fig:final_exp} shows that \pname{} often reaches the ProphetKV-quality region at lower TTFT, and in several model--task panels reaches full-prefill-level quality while remaining left of full prefill. Figure~\ref{fig:benchmark_aggregate_ttft_quality} confirms the same trend in aggregate. The reason is that \pname{} obtains query awareness through chunk-anchor probing and critical-layer localization, so selection fits the layer-wise cache-fusion pipeline instead of becoming a blocking full-view stage. \textbf{The TTFT result shows that user-query-aware selection must be pipeline-compatible to improve TTFT.}

\subsubsection{Recomputation-Ratio Sweep}
\begin{figure}[t]
  \centering
  \includegraphics[width=\linewidth]{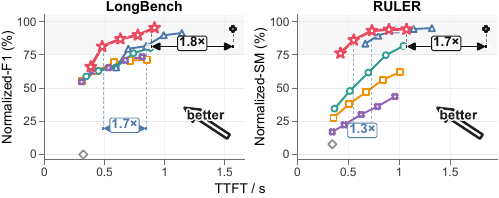}
  \caption{\pname{} preserves quality while reducing TTFT by 1.7$\times$ over full prefill and 1.5$\times$ over ProphetKV.}
  \Description{Two aggregate quality versus time-to-first-token plots for LongBench and RULER. The plots summarize the fine-grained model-by-dataset results and show QCFuse reaching comparable quality at lower TTFT than the competing baselines.}
  \label{fig:benchmark_aggregate_ttft_quality}
\end{figure}

Increasing $\rho$ recomputes more retrieved-context token positions, so TTFT rises and quality usually moves closer to full prefill. The main exception in Figure~\ref{fig:final_exp} is Mistral-v0.3-7B on \texttt{vt}, where direct reuse is already above full prefill and more recomputation pulls behavior back toward the full-prefill reference. At $\rho=0.5$, \pname{} reaches the full-prefill-quality region while recomputing only half of the retrieved-context positions. In the aggregate view in Figure~\ref{fig:benchmark_aggregate_ttft_quality}, averaging the matched-quality operating points gives a 1.7$\times$ TTFT speedup over full prefill and a 1.5$\times$ speedup over ProphetKV.

Overall, \pname{} preserves full-prefill-level quality while avoiding ProphetKV's blocking full-view selector. \textbf{The end-to-end takeaway is that \pname{} shifts the quality--TTFT frontier by reaching full-prefill-level quality with partial recomputation and lower TTFT.}

\subsection{Long-Context Stress Test}
\label{sec:exp_long_context}

This experiment tests whether \pname{} keeps its quality advantage as retrieved context grows. We fix $\rho=0.5$ because the end-to-end sweep shows that this recomputation ratio already reaches the full-prefill-quality region; fixing it isolates the effect of context length rather than mixing length changes with a larger recomputation budget. We then vary RULER context length by increasing either the number of chunks, which adds more retrieved units and distractors, or the chunk size, which makes each reused chunk longer and harder to localize within.

\begin{figure*}[t]
  \centering
  \includegraphics[width=\textwidth]{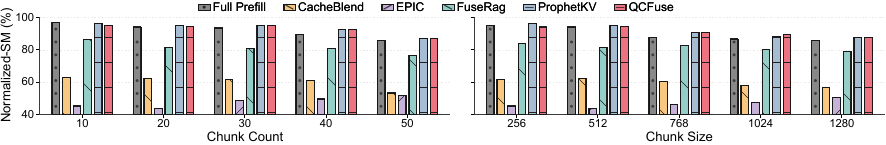}
  \caption{\pname{} maintains high Normalized-SM as RULER contexts grow. The sweep varies chunk count and chunk size at fixed $\rho=0.5$.}
  \Description{A bar chart summarizing Normalized-SM on RULER as context length increases by varying chunk count and chunk length.}
  \label{fig:ruler_context_sweep}
\end{figure*}

Figure~\ref{fig:ruler_context_sweep} shows that \pname{} stays in the high-quality group across both sweeps and remains consistently stronger than the query-agnostic or lightweight query-aware baselines. At shorter contexts, full prefill and ProphetKV can retain a small advantage because there are fewer redundant tokens and each missed token matters more. Across the sweep, CacheBlend, EPIC, and FusionRAG remain in a lower-quality region and stay noticeably below full prefill, while \pname{} remains close to the full-prefill/ProphetKV group. The advantage of user-query-aware allocation becomes clearer in this regime because longer inputs contain more distractors and repeated evidence, so selecting answer-relevant regions matters more than spending the recomputation budget by request-agnostic or weakly query-aware rules. \textbf{\pname{} remains stable under longer contexts because its compact selector uses the user query to spend a fixed recomputation budget on useful tokens.}

\subsection{Bandwidth-Constrained Cache Loading}
\label{sec:exp_bandwidth}

This experiment tests whether query-aware selection becomes an I/O bottleneck when cache-loading bandwidth drops. We fix $\rho=0.5$ for a fair TTFT comparison because \pname{} and ProphetKV are already in a similar quality region at this ratio, so the remaining difference mainly reflects selection and cache-loading overhead rather than different answer quality. We compare full prefill, ProphetKV, and \pname{} because ProphetKV is the strongest full-context query-aware selector, while full prefill provides the no-cache-loading reference.

\begin{figure}[t]
  \centering
  \includegraphics[width=\linewidth]{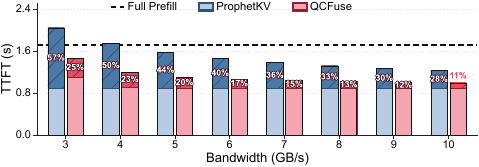}
  \caption{\pname{} is less sensitive to cache-loading bandwidth than ProphetKV. The highlighted \emph{select in prefill} segment stays small for \pname{}.}
  \Description{A bandwidth-sensitivity plot comparing QCFuse and ProphetKV TTFT under different NVMe bandwidth settings, with the select-in-prefill portion marked.}
  \label{fig:bandwidth_constrained_ttft}
\end{figure}

Figure~\ref{fig:bandwidth_constrained_ttft} highlights the prefill-side selection cost marked as \emph{select in prefill}. At high bandwidth, ProphetKV can hide part of this cost, so the TTFT gap is moderate. As bandwidth drops, however, its \emph{select in prefill} segment grows quickly because ProphetKV must load broad KV-cache evidence before the selected tokens are known; around 3 GB/s, this blocking selection path can make it slower than full prefill. \pname{} loads only chunk anchors and critical-layer keys before scoring tokens, so its \emph{select in prefill} time remains much smaller. The advantage is therefore more pronounced at low transfer speeds because \pname{} reduces the serialized selection time, not just the recomputation work. \textbf{Under bandwidth pressure, \pname{} keeps query-aware selection useful by making the selector small enough to avoid blocking cache-fusion execution.}

\subsection{Serving Throughput}
\label{sec:exp_throughput}

This stress test asks whether the single-request TTFT gains remain as the average request rate increases. We fix $\rho=0.5$ for a fair throughput comparison because \pname{} and ProphetKV operate in a similar quality region at this ratio, so differences in the TTFT--throughput curve mainly reflect serving-path overhead rather than answer-quality tradeoffs. We compare full prefill, ProphetKV, and \pname{} because full prefill is the full-context reference, while ProphetKV is the strongest full-context query-aware selector.

\begin{figure*}[t]
  \centering
  \includegraphics[width=\textwidth]{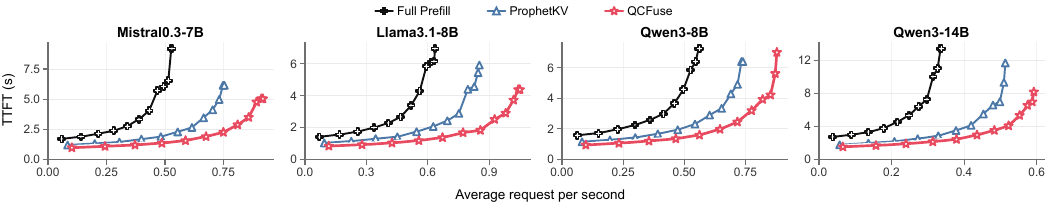}
  \caption{\pname{} sustains lower TTFT at higher request throughput across models.}
  \Description{A throughput plot comparing full prefill, ProphetKV, and QCFuse across four models. The x-axis reports request throughput and the y-axis reports time to first token.}
  \label{fig:ttft_by_model}
\end{figure*}

Figure~\ref{fig:ttft_by_model} shows that \pname{} shifts the TTFT--throughput curve to the right on all four models. At a comparable TTFT, the serving stack can admit more requests, and at a comparable request rate it returns the first token earlier. The curve shape is also important. Full prefill reaches the steep-TTFT region earliest because every request processes the whole retrieved context, while ProphetKV reduces recomputation but still adds a blocking cache-loading and scoring stage before recomputation can start. \pname{} keeps this prefill-side selection stage compact, so its TTFT increases more slowly as request pressure grows. \textbf{\pname{} sustains lower TTFT under increasing request load because it removes serialized selection work before recomputation starts.}

\subsection{Component and Design Ablation}
\label{sec:exp_ablation}

This ablation checks whether \pname{}'s two compressed-view components improve selection quality across the recomputation-ratio sweep. The stress tests fix $\rho=0.5$ to compare systems at a representative quality-matched operating point, whereas this ablation averages quality over $\rho=0.1$--$0.5$ to measure each component's contribution across the selection-budget sweep.

\subsubsection{Chunk Anchors}
\begin{figure}[t]
  \centering
  \includegraphics[width=\linewidth]{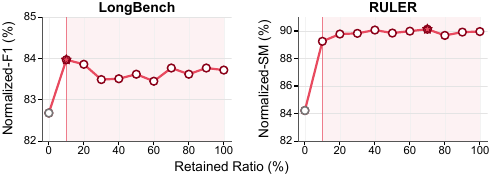}
  \caption{A small chunk-anchor set captures most of the average quality gain over $\rho=0.1$--$0.5$.}
  \Description{An anchor-ratio ablation plot for LongBench and RULER. Quality improves sharply from zero anchor ratio to a small anchor ratio, then changes more gradually as the anchor ratio increases.}
  \label{fig:zip_ratio_ablation}
  \vspace{-1.5em}
\end{figure}

Figure~\ref{fig:zip_ratio_ablation} varies the anchor retained ratio $r_a$. The $r_a=0$ setting removes chunk anchors and leaves the selector with user-query-only probing, while larger $r_a$ values expose more cached chunk evidence. The largest change occurs from $r_a=0$ to $r_a=0.1$. Without anchors, the user-query hidden states mostly encode the question wording and cannot reliably identify which retrieved tokens should be recomputed. A small anchor set is enough to make the query tokens context-aware.

The benchmark-level trends are different but consistent with the task structure. LongBench reaches its best region near $r_a=0.1$, because semantic QA mainly needs the selector to locate relevant evidence regions. RULER can benefit slightly from more anchors because exact retrieval and tracking tasks depend more on repeated keys, exact values, and variable bindings. In both cases, the extra gain beyond a small anchor set is much smaller than the gain from adding anchors at all. \textbf{Chunk anchors help because they add compact retrieved-context evidence, and most of the benefit comes from the first small anchor budget.}

\subsubsection{Critical Layers}
\begin{figure}[t]
  \centering
  \includegraphics[width=\linewidth]{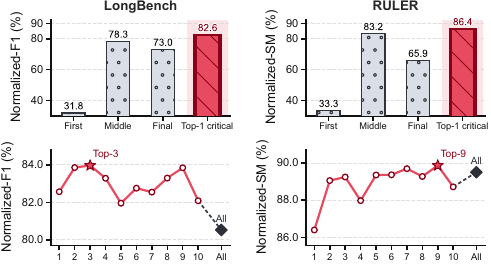}
  \caption{The profiled Top-1 critical layer gives the highest average quality over $\rho=0.1$--$0.5$.}
  \Description{A single-column ablation plot summarizing how different layer-selection choices affect QCFuse quality and time-to-first-token.}
  \label{fig:layer_ablation_summary}
  \vspace{-1em}
\end{figure}

Figure~\ref{fig:layer_ablation_summary} varies the layer view used for scoring, comparing the first layer, the fixed middle layer at $\lfloor L/2 \rfloor$, the final layer, all-layer aggregation, and the profiled Top-1 critical layer. All-layer aggregation is included as an informative but non-pipeline-friendly reference. Shallow layers are too lexical and local, so they miss multi-hop or cross-chunk relevance. Final layers are more semantic, but they can mix evidence localization with generation behavior, output formatting, and attention sinks. All-layer visibility is informative but expensive, and risks recreating ProphetKV's blocking full-view selection path.

The profiled Top-1 critical layer achieves the highest average quality on both LongBench and RULER, while the fixed middle layer is the strongest or near-strongest fixed-layer alternative. This supports the default Top-3 critical-layer set in Section~\ref{sec:critical_layer}, since the useful localization signal concentrates in profiled middle layers rather than in the earliest layer, the final layer, or a full all-layer scan. The final-layer result also avoids a contradiction with final-layer baselines. A final-layer-only signal can be useful, but it is not the best localization view in this ablation. \textbf{Critical-layer selection provides enough layer visibility for token localization without paying for a full-layer selector.}

\subsection{Summary of Findings}
\label{sec:exp_summary}

The experiments give a consistent picture of where the gain comes from. In the end-to-end sweep, \pname{} reaches the full-prefill-quality region with partial recomputation and lower TTFT than full prefill and ProphetKV. The stress tests then examine the same operating point under harder serving conditions, including longer RULER contexts, reduced cache-loading bandwidth, and increasing request load. In all three cases, \pname{} keeps the prefill-side selection step short by using compact query-conditioned chunk evidence and a small critical-layer view before recomputation. The ablations explain this behavior. Chunk anchors supply enough retrieved-context signal for query-aware token selection, while profiled critical layers provide the needed depth view without all-layer scanning. \textbf{Overall, \pname{} improves the quality--TTFT frontier because its selector is query-aware, compact, and pipeline-compatible.}

\section{Conclusion}
\label{sec:discussion}

\pname{} is a compressed-view query-aware selector for efficient RAG KV cache fusion, motivated by the observation that selective recomputation needs token-level evidence conditioning and layer-level localization without stalling the layer-wise cache-fusion pipeline.
\pname{} combines chunk-anchor query probing with critical-layer profiling to obtain compact query-aware evidence for selecting tokens to recompute.
Implemented in SGLang with a pipelined cache-fusion runtime, \pname{} reaches full-prefill-level quality while reducing prefill-stage latency compared with full prefill and strong cache-fusion baselines.


\bibliographystyle{ACM-Reference-Format}
\bibliography{sample}

\end{document}